\documentclass[11pt, a4paper, twocolumn]{article}

\usepackage{color}

\usepackage[utf8]{inputenc}
\usepackage[english]{babel}

\usepackage[top=2cm, bottom=2cm, left=1.9cm, right=1.9cm]{geometry}

\usepackage{titlesec}
\titlespacing{\section}{0pt}{\parskip}{\parskip}
\titlespacing{\subsection}{0pt}{\parskip}{\parskip}
\titlespacing{\subsubsection}{0pt}{\parskip}{\parskip}

\usepackage{indentfirst}
\usepackage[export]{adjustbox}
\usepackage{graphicx}
\usepackage{amssymb}

\usepackage[dvipsnames]{xcolor}
\graphicspath{{figure/}{table/}}

\usepackage[figurename=Fig.]{caption}
\usepackage{caption}
\usepackage[colorlinks, citecolor=ForestGreen]{hyperref}

\usepackage{cite}

\usepackage{abstract}
\usepackage{multirow}
\usepackage{array}
\newcolumntype{*}{>{\global\let\currentrowstyle\relax}}
\newcolumntype{^}{>{\currentrowstyle}}
\newcommand{\rowstyle}[1]{\gdef\currentrowstyle{#1}%
	#1\ignorespaces
}

\title{\textbf{Foreground Segmentation Using a Triplet Convolutional Neural Network for Multiscale Feature Encoding}\footnote{Note that this paper is under consideration at Pattern Recognition Letters.}}
\date{\vspace*{2pt}}
\author{
	\normalsize \textbf{Long Ang Lim}\\
	\normalsize Ankara University\\
	\normalsize Department of Computer Engineering\\
	\normalsize lim.longang@gmail.com
	\and
	\normalsize \textbf{Hacer Yalim Keles}\\
	\normalsize Ankara University\\
	\normalsize Department of Computer Engineering\\
	\normalsize hkeles@ankara.edu.tr
}

\begin{document}
\maketitle

\begin{abstract}{
	\vspace*{-1.5em}
	\it A common approach for moving objects segmentation in a scene is to perform a background subtraction. Several methods have been proposed in this domain. However, they lack the ability of handling various difficult scenarios such as illumination changes, background or camera motion, camouflage effect, shadow etc. To address these issues, we propose a robust and flexible encoder-decoder type neural network based approach. We adapt a  pre-trained convolutional network, i.e. VGG-16 Net, under a triplet framework in the encoder part to embed an image in multiple scales into the feature space and use a transposed convolutional network in the decoder part to learn a mapping from feature space to image space. We train this network end-to-end by using only a few training samples. Our network takes an RGB image in three different scales and produces a foreground segmentation probability mask for the corresponding image. In order to evaluate our model, we entered the Change Detection 2014 Challenge (changedetection.net) and our method outperformed all the existing state-of-the-art methods by an average F-Measure of 0.9770. Our source code will be made publicly available at \href{https://github.com/lim-anggun/FgSegNet}{https://github.com/lim-anggun/FgSegNet}.
	
}\end{abstract}


\section{Introduction}
	Moving objects segmentation from video sequences that are captured from stationary/non-stationary cameras is a crucial computer vision problem for efficient video surveillance \cite{brutzer2011evaluation}, human tracking \cite{porikli2003human}, action recognition \cite{zhu2015human,poppe2010survey}, traffic monitoring \cite{cheung2004robust}, motion estimation and anomaly detection applications \cite{basharat2008learning}. A common method for segmenting moving objects in a scene is to perform a background subtraction, in which moving objects are considered as foreground pixels and non-moving objects are considered as background pixels. This binary classification problem has been extensively studied and improved over the years, and several approaches have been proposed concurrently \cite{stauffer1999adaptive,zivkovic2004improved,barnich2011vibe,van2012background,st2015subsense,braham2016deep,babaee2017deep,WANG201766,sakkos2017end}. There are many challenges in developing a robust background subtraction algorithm: sudden or gradual illumination changes, shadows cast by foreground objects, dynamic background motion (waving tree, rain, snow, air turbulence), camera motion (camera jittering, camera panning-tilting-zooming), camouflage or subtle regions, i.e. similarity between foreground pixels and background pixels. However, conventional approaches only perform well on some specific type of scenarios, and lack the capability to handle the problem in a general setting. Consider the traffic monitoring and video surveillance domains; the approach should segment the moving objects in a robust way under various weather conditions and aforementioned challenges independently from the positioning of the camera.
	
	Convolutional neural networks (CNNs) have recently been very popular and have been used successfully in object recognition \cite{krizhevsky2012imagenet,zeiler2014visualizing,szegedy2015going,he2016deep}, scene labeling \cite{farabet2013learning,long2015fully,pinheiro2014recurrent} and in many other domains \cite{lai2015recurrent,venugopalan2014translating,vinyals2015show,xu2015show,karpathy2015deep}. They are very powerful in extracting low-, mid- and high-level feature representations from images which turn out to be useful in various computer vision problems including the foreground/background segmentation.
	
	In this work, we propose a robust, and flexible approach for moving objects segmentation using a triplet CNN and a transposed convolutional neural network (TCNN) attached at the end of it in an encoder-decoder structure. We adapt the first four blocks of the pre-trained VGG-16 Net \cite{simonyan2014very} at the beginning of our CNNs under a triplet framework as our multiscale feature encoder and integrate a novel decoder network at the end of it to map the features to a pixel-level foreground probability map. We then apply thresholding to this map to obtain binary segmentation labels. To the best of our knowledge, this is the first work that applies this technique in the moving object segmentation problem. The proposed solution is simple compared to the previous approaches, yet produce impressive segmentation results. We evaluated our method with the largest publicly available CDnet2014 dataset \cite{wang2014cdnet}, which contains pixel-level ground truths; the test results reveal that our method significantly improves the previous best method in terms of average F-Measure and average MCC across 11 categories (Table \ref{table:6}). We will call our Foreground Segmentation Network shortly as FgSegNet from this on.
	
	The rest of this paper is organized as follows. A brief summary about previous works is described in Section \hyperref[sec:2]{2}, we introduce our approach in Section \hyperref[sec:3]{3}, we report our experiment results in Section \hyperref[sec:4]{4}, and conclusion and future work in Section \hyperref[sec:5]{5}.
	
\section{Related Works}
\label{sec:2}
	In the past several years, various methods have been proposed in foreground objects segmentation problem. This problem can be restated as determining a foreground mask from an image sequence where the masked regions refer to the moving objects in the scene. In order to extract a foreground mask from a specific scene, one should build a robust and flexible background model which can be utilized in each frame of an image sequence to determine foreground regions of that scene.
	
	In a classical background subtraction method, a given static frame or the previous frame is utilized as the background model. Although intuitively correct, this method is very sensitive to dynamic changes in the background. To model the variance in the background model more effectively, probabilistic approaches are adapted; one of the most widely used probabilistic model is the Gaussian Mixture Model (GMM) \cite{stauffer1999adaptive}. Stauffer and Grimson use a mixture of Gaussians to model each pixel as a background pixel or a foreground pixel instead of modeling all pixel values as one distribution. In \cite{kaewtrakulpong2002improved}, Kaewtrakulpong and Bowden modified the update equation of \cite{stauffer1999adaptive} in order to improve accuracy of segmentation and proposed a shadow detection scheme to eliminate shadow using an existing GMM. In \cite{zivkovic2004improved}, Zivkovic improved the GMM algorithm by constantly adapting the number of Gaussian distribution for each pixel; in contrast to \cite{zivkovic2004improved},\cite{stauffer1999adaptive} used a fixed number of Gaussian distributions.
	
	As for non-parametric approaches, in \cite{barnich2011vibe}, Barnich and Van Droogenbroeck proposed a pixel-based method, which is called ViBe, where current pixel value is compared to its closest sample within the collection of samples. This method is robust against small camera movements and noise. Van Droogenbroeck and Paquot \cite{van2012background} extensively studied and proposed several modifications to the original work of ViBe by adjusting some parameters. St-Charles et al. \cite{st2015subsense} also proposed a non-parametric method to combine color intensities and Local Binary String Pattern (LBSP) which is capable of detecting camouflaged objects and handling illumination variations. They also proposed a method based on a word-based model, called PAWCS \cite{st2015self}. By using color and texture information, the appearance of pixels are registered as background words which are considered as good representational models when they are persistent. Bianco et al. \cite{bianco2017far} proposed to use a Genetic Programming to select best approaches from existing change detection methods and combined them, they applied post processing technique to determine final labels.
	\renewcommand{\thefootnote}{\fnsymbol{footnote}}
	\footnotetext[2]{The values are obtained from their paper.}
	
	\begin{figure*}[!ht]
		\includegraphics[width=\textwidth, center]{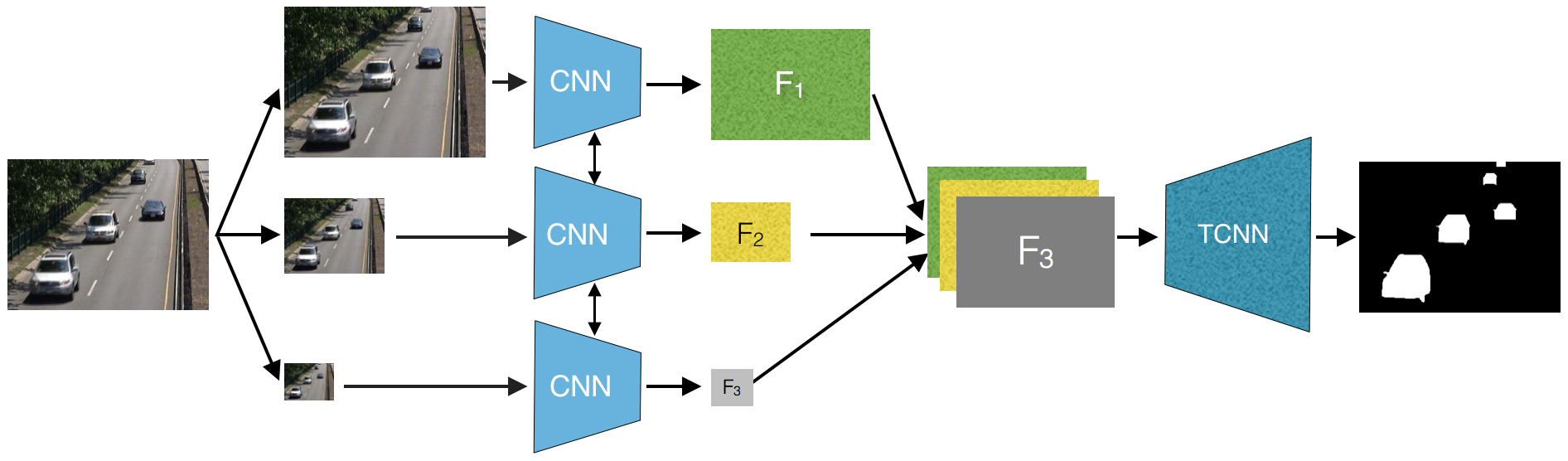}
		\caption{The FgSegNet Architecture}
		\centering
		\label{fig:fig1}
	\end{figure*}
	Recently, deep learning based approaches have been proposed by many researchers that are based on learning the hidden features in scenes and segmenting foreground objects in video sequences using these features. In \cite{braham2016deep}, Braham and Van Droogenbroeck proposed a scene specific method using CNNs. More precisely, a single background model was built for a specific scene. For each frame in a video sequence, image patches that are centered on each pixel are extracted and then they are combined with corresponding patches from the background model. After that these combined patches are fed to the network to predict probability of foreground pixels. They used half of the training examples for training their network (by considering the range of the frames that contain ground truth labels) and took the remaining frames for testing. This method achieved an average F-Measure of 0.9046\textcolor{magenta}{\footnotemark[2]} on CDnet2014 dataset. Their method is computationally expensive due to large number of patches extracted from each frame. Conversely, in \cite{WANG201766}, Wang et al. proposed an image-wise method without using any background models. They trained scene specific networks using 200 frames by manual selection and have an overall F-Measure of 0.95\textcolor{magenta}{\footnotemark[2]} in CDnet2014 dataset. Instead of training a network for a specific scene, Babaee et al. \cite{babaee2017deep} trained their model all at once by combining training frames from various video sequences; in particular, including 5\% of frames from each video sequence. They followed the same training procedure as in \cite{braham2016deep}, in which image-patches were combined with background-patches then fed to the network. They obtained an F-Measure of 0.7548\textcolor{magenta}{\footnotemark[2]}. Recently, Sakkos et al. \cite{sakkos2017end} used a 3D convolution technique to track temporal changes in video sequences, without using any background models in training. Their approach performed with an average F-Measure of 0.9507\textcolor{magenta}{\footnotemark[2]} in CDnet2014 dataset.
	
	In this work, we generated scene specific models using only a few frames, i.e. 50 and 200, similar to Wang et al. \cite{WANG201766}. Using the same methodology with theirs in the training frame selection, our model outperformed all the reported methods by an overall F-Measure of 0.9770 and ranked as number one in CDnet 2014 Challenge.
	
\section{The Method}
\label{sec:3}
	In this section, we clarify the details of our approach in three separate sections: (1) training examples selection, (2) network architecture and (3) implementation details.
	
	\subsection{Training Examples Selection}
	\label{sec:3.1}
		Selection of the frames for training scene specific models can be crucial and may require attention if the background is dynamic and the images in the scene contain important artifacts such as \textit{thermal}, \textit{dynamicBackground}, \textit{badWeather}, or \textit{turbulence} categories in CDnet2014 dataset. For a static video sequence which has less background motion, such as slightly waving trees, only a number of training examples, i.e. 50 frames, will be sufficient. The frames can be selected randomly by focusing more on the frames that contain some foreground objects. This strategy helps the network to learn and segment foreground pixels more accurately. However, for more complex scenes and dynamic backgrounds or camera panning-tilting-zooming video sequences, it is better to select more training examples, i.e. 200 frames, by including different parts of the scene in the selected examples. The content of the training frames may include foreground or background parts or both. One can select \textit{n} number of frames, where \textit{n $\ll$ N} and \textit{N} is the total number of frames in a video sequence. In our experiments, we manually and randomly selected 50 and 200 frames for two separate trainings. Next section, we discuss the imbalanced class sample problem in supervised binary classification, which needs attention to generate robust models in this domain.
		
		\subsubsection{Working with imbalanced data}
		\label{sec:3.1.1}
			In a supervised training setting, the imbalanced number of training examples for different class categories may cause bias problems in classification; this is an active research problem \cite{he2009learning,chawla2004special}. It is also the case in foreground object segmentation training since the distribution of the background pixels usually out-weights the set of foreground pixels in wide field-of-view surveillance camera settings. In this binary classification domain, it may appear in severe ratios of 100:1, 10000:1 or even 10000:0, which causes sub-optimal performance in classification. We can alleviate this issue in two levels: data level and algorithmic level. In our problem, dealing with imbalanced classes in data level is difficult, hence we choose to alleviate it in the algorithmic level. We implemented this by penalizing the computed loss more if a foreground pixel is classified as a background pixel. We compute the class-penalization-weights by using foreground/background pixels distribution for each training frame using the ground-truth, independently. In particular, in the video sequences of CDnet2014 dataset, background pixels are severely more than foreground pixels, hence we applied the weighted loss computation during training.
			
			In foreground segmentation (or background subtraction) problem, F-Measure (or F1-Score) is widely used to evaluate model performances. However, as it is also claimed by \cite{boughorbel2017optimal}, F-Measure is sensitive to imbalanced data since it does not incorporate true negatives into account. In \cite{boughorbel2017optimal}, it is argued that MCC \cite{matthews1975comparison} is more suitable in imbalanced classes classification problems due to the incorporation of true negatives into consideration. In this paper, we report both F-Measure and MCC performances of our method in Section \ref{sec:4.1}.

	\subsection{FgSegNet Network Architecture}
	\label{sec:3.2}
		Our end-to-end network architecture, which contains a triplet CNN that operates in three different scales for feature encoding and a transposed convolutional network for decoding, is depicted in Fig. \ref{fig:fig1}. This network learns a function \textit{\textbf{f}} that maps a given set of raw pixel values $\mathit{P^{R}}$ to a set of probability values, i.e. values between 0 and 1, that represent the foreground probability map $\mathit{P^{M}}$, defined by \textit{\textbf{f}}:$\mathit{P^{R}}\rightarrow\mathit{P^{M}}$. To correctly learn this mapping (\textit{\textbf{f}}), contextual information around the neighborhood of each pixel is essential. Learning to classify a pixel from a small fixed window, which is centered on it, is difficult. More precisely, consider classifying a very small region of a big cat that is flat, i.e. sharing similar intensities all around it; it is hard to tell whether it is a part of the cat or not by analyzing this local content without context. 
		\begin{table}[h]
			\captionsetup{width=\columnwidth}
			\renewcommand{\arraystretch}{1.25}
			\caption{ Our network configuration. A modified CNNs (VGG16) is from block 1 to 4, where block 0 is a RGB input image. TCNN is from block 5 to 9, where block 9 is the output probability mask from the network. Note that ReLU non-linearities are applied after every convolutional and transposed convolutional layers, except the last transposed convolutional layer which we apply sigmoid activation (for briefing, we do not show here).}
			\label{table:1}
			\scriptsize
			\begin{tabular}{|p{1mm}|p{1.5cm}p{1.5cm}|p{1mm}|p{1.5cm}p{1.5cm}|}
				\hline
				&\multicolumn{2}{c|}{CNNs (VGG-16)}&&\multicolumn{2}{c|}{TCNN}\\
				\hline
				0&WxHx3&rgb image&9&WxHx1&F=1x1,S=1, seg.mask\\
				
				\hline
				
				\multirow{3}{*}{1}&WxHx64&F=3x3,S=1&\multirow{3}{*}{8}&&\\
				&WxHx64&F=3x3,S=1&&WxHx64&F=5x5,S=2\\
				&max-pool.&F=2x2,S=2&&& \\
				\hline
				\multirow{3}{*}{2}&$\frac{W}{2}$x$\frac{H}{2}$x128&F=3x3,S=1&\multirow{4}{*}{7}&$\frac{W}{2}$x$\frac{H}{2}$x128&F=1x1,S=1\\
				&$\frac{W}{2}$x$\frac{H}{2}$x128&F=3x3,S=1&&$\frac{W}{2}$x$\frac{H}{2}$x64&F=3x3,S=1\\
				&max-pool.&F=2x2,S=2&&$\frac{W}{2}$x$\frac{H}{2}$x64&F=1x1,S=1\\
				\hline
				\multirow{3}{*}{3}&$\frac{W}{4}$x$\frac{H}{4}$x256&F=3x3,S=1&\multirow{3}{*}{6}&$\frac{W}{2}$x$\frac{H}{2}$x256&F=1x1,S=1\\
				&$\frac{W}{4}$x$\frac{H}{4}$x256&F=3x3,S=1&&$\frac{W}{2}$x$\frac{H}{2}$x64&F=5x5,S=2\\ 
				&$\frac{W}{4}$x$\frac{H}{4}$x256&F=3x3,S=1&&$\frac{W}{4}$x$\frac{H}{4}$x64&F=1x1,S=1\\ \cline{1-6}
				
				\multirow{6}{*}{4}&$\frac{W}{4}$x$\frac{H}{4}$x512&F=3x3,S=1&&&\\
				&dropout&rate=0.5&&&\\
				&$\frac{W}{4}$x$\frac{H}{4}$x512&F=3x3,S=1&\multirow{3}{*}{5}&$\frac{W}{4}$x$\frac{H}{4}$x512&F=1x1,S=1\\
				&dropout&rate=0.5&&$\frac{W}{4}$x$\frac{H}{4}$x64&F=3x3,S=1\\
				&$\frac{W}{4}$x$\frac{H}{4}$x512&F=3x3,S=1&&$\frac{W}{4}$x$\frac{H}{4}$x64&F=1x1,S=1\\
				&dropout&rate=0.5&&&\\\cline{1-6}
			\end{tabular}
		\end{table}
		To understand the contextual relation of this local region with its surrounding, the network needs to engage global information in multiple scales.
		
		These ideas inspired us to use full-size and multi-scale images in the network training.  A fixed receptive field is operated on these scales. As demonstrated in Fig. \ref{fig:fig1}, we downsample the input image, which is represented in RBG color space, by a factor of two using a Gaussian pyramid with a sigma shown below:
		\begin{equation}
		sigma = \frac{downscale}{3}
		\end{equation} 
		where the \textit{downscale} is the downscale factor, which we set to 2 in our implementation. More precisely, given an input image \textit{I}, it is downscaled into $I_{i}$:$i\in[0,1,2]$  where $I_{0}$ is the original size of the image. These three images are fed simultaneously to our triplet CNN in parallel. Note that the architecture of the CNNs in the triplet are exactly the same and they share weights (refer to Section \ref{sec:3.2.1} for the architectural details). The resultant embeddings of each input is denoted by $F_{i}$:$i\in[1,2,3]$, where $F_{1}$, $F_{2}$ and $F_{3}$ are the embeddings of the inputs $I_{0}$, $I_{1}$ and $I_{2}$, respectively. These feature embeddings are then re-arranged to compose the combined feature representation of the decoding network. In this context, $F_{2}$ and $F_{3}$ are upscaled to match the scale of $F_{1}$, and then concatenated along their depth axis to form the combined feature map, i.e. \textit{F}. Finally, \textit{F} is fed into a single TCNN to learn the weights for decoding. Final output is a segmentation mask that has the same size as the original input ($I_{0}$). The details of our encoding and decoding network configurations are provided below.
		
		\subsubsection{The Triplet CNN Configuration}
		\label{sec:3.2.1}
			CNNs perform well, and even outperform human performance by some margins, in various problems in different domains. To gain a deeper insight about what CNNs learn, we can inspect the visualizations of the filters that are learnt at each layer \cite{zeiler2014visualizing}. These visualizations show that the lower layers learn some generic low-level features such as color blobs, edges in various directions, textures which are useful in many tasks when used as feature representations. Motivated by this generic feature encoding properties of the CNNs, we utilize a triplet CNN that contains three copies of a CNN that operate in parallel with the same input in three different scales. The first four blocks of these networks are modified copies of the pre-trained VGG-16 Net \cite{simonyan2014very}; we removed the third and fourth max pooling layers and insert dropouts between each layer of fourth convolutional block as illustrated in Fig. \ref{fig:fig2} (for full network architecture of VGG-16 Net, one may refer to the original paper in \cite{simonyan2014very}).
			
			The input to each CNN is raw RGB images in different sizes. Assuming that the input image size is WxHx3, where W is the image width, H is the image height and 3 is the RGB color channels, it is transformed to 64 feature maps of size WxH at the end of the first convolutional block, then these feature maps are downsampled by a 2x2 max pooling layer with a stride 2 and transformed into 128 feature maps of size $\frac{W}{2}\times\frac{H}{2}$ at the end of the second block. Again, these feature maps are downsampled by a 2x2 max pooling layer with a stride 2 and transformed into 256 feature maps of size $\frac{W}{4}\times\frac{H}{4}$ at the end of the third block. Finally, these feature maps are transformed into 512 feature maps of size $\frac{W}{4}\times\frac{H}{4}$ at the end of the fourth block.
			
			In our segmentation approach, we use only a few training examples for model generation; hence, to avoid overfitting, we apply dropout regularization after each convolutional layer in the fourth convolutional block. Note that zero padding is applied in all the convolutional layers in our network to preserve spatial dimensions of the inputs in the outputs. The details of the encoding network configuration are presented in Table \ref{table:1}.
			\begin{center}
				\includegraphics[width=0.9\columnwidth, center]{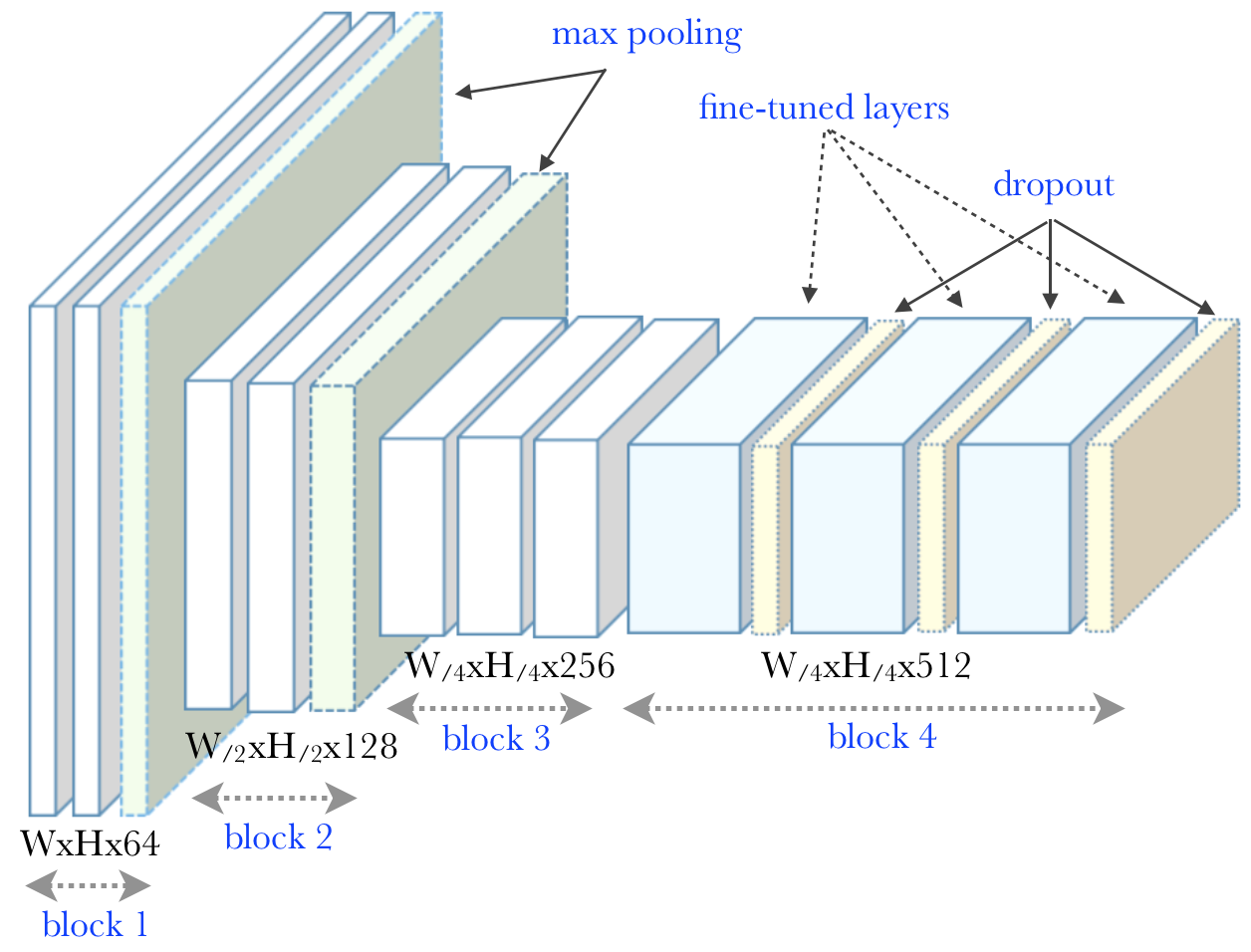}
				\captionsetup{width=0.9\columnwidth}
				\captionof{figure}{The architecture of each CNN in the triplet network.}
				\label{fig:fig2}
			\end{center}
		\subsubsection{TCNN Configuration}
		\label{sec:3.2.2}
			The output of the encoding network, i.e. \textit{F}, is a concatenated form of the feature maps in three different scales. This map is fed to the TCNN to learn the weights for decoding the feature maps; the output will be a dense probability mask (Fig. \ref{fig:fig3}). In our network, \textit{F} has a large depth, i.e. 1536, due to concatenation of features across three different scales. For computational efficiency and to increase non-linearity of the decision function in our network, we use 1x1 transposed convolutional layers in each block to project a high dimensional feature map depth into a lower dimension.
		
			If we consider block 5 in TCNN, which is specified in detail in 
			\begin{center}
				\includegraphics[width=0.9\columnwidth, center]{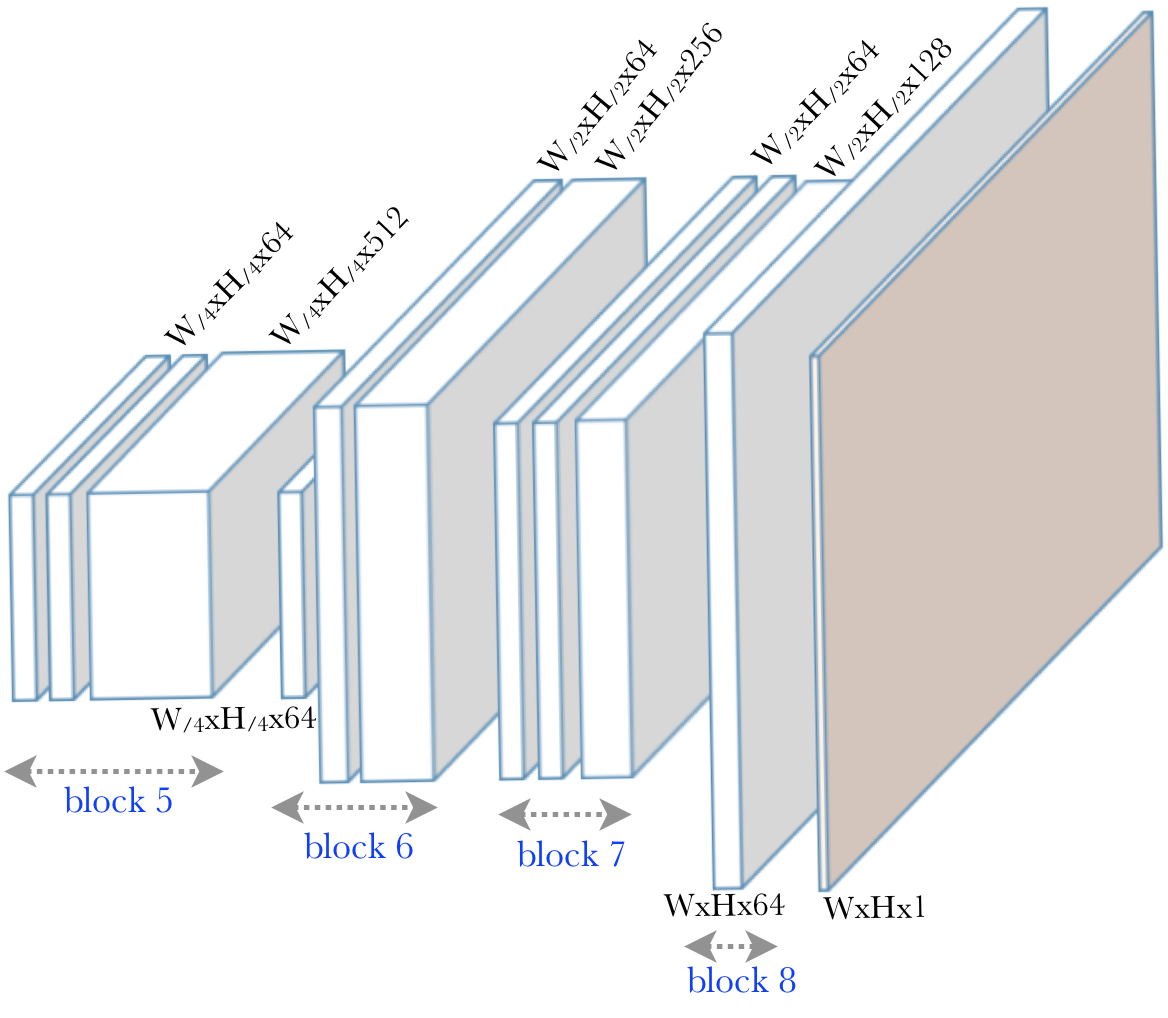}
				\captionsetup{width=0.9\columnwidth}
				\captionof{figure}{The TCNN architecture}
				\label{fig:fig3}
			\end{center}
			the bottom-right row of Table \ref{table:1}, the concatenated feature \textit{F} of shape $\frac{W}{4}\times\frac{H}{4}\times1536$ is projected into $\frac{W}{4}\times\frac{H}{4}\times64$ using a 1x1 transposed convolution with a stride of 1. The projected features are operated with 3x3 transposed convolution, with a stride of 1 and projected into $\frac{W}{4}\times\frac{H}{4}\times64$. Finally, it is projected into $\frac{W}{4}\times\frac{H}{4}\times512$ to enlarge the number of feature maps along the depth axis. The similar structure of layers are operated in blocks 6 and 7, except that we apply 5x5 transposed convolution with a stride of 2 to upscale feature maps by a factor of two in block 6. Moreover, we reduce the number of feature maps to 256 and 128 for block 6 and 7, respectively. In block 8, we operate 5x5 transposed convolution with a stride of 2 to enlarge feature maps to match the original size of the input image. In block 9, we project 64 feature maps of block 8 into 1 feature map by operating a 1x1 transposed convolution with a stride of 1. Finally, a sigmoid function is applied to the last layer to generate a probability mask for each pixel to encode the probability of being a foreground pixel by a value that is between 0 and 1. Next section demonstrates the implementation details of our approach.
			
	\subsection{Implementation Details}
		We perform our implementation using Keras framework \cite{chollet2017keras} with Tensorflow backend with an NVIDIA GTX 970 GPU. As we describe in the previous section, our modified VGG-16 Net contains a total of 10 convolutional layers, 2 max pooling layers and 3 dropout layers. TCNN contains 11 transposed convolutional layers, where the last transposed convolutional layer is the result of projecting 64 feature maps into a grayscale image. In this part we do not apply any unpooling. Note that ReLU non-linearities are applied to every (transposed) convolutional layers in both modified VGG-16 Net and TCNN, except the last transposed convolutional layer where a sigmoid activation is used to predict a probability mask. Besides dropout, to alleviate overfitting in our network, we apply L2 regularization to the weights in the first transposed convolutional layers, in the blocks 5, 6, 7 and 8.
		
		\subsubsection{Training Details}
			As illustrated in Fig. \ref{fig:fig2} and Table \ref{table:1}, during training we adapted the weights of convolutional block 4 and kept the weights of convolutional blocks 1, 2 and 3, as they are in the initial VGG-16 Net model. We train our model using \textit{N} training examples, where \textit{N} is fixed and 50 (or 200), in our experiments. Each example contains \textit{M} pixels denoted by:
			\begin{equation}
				\resizebox{0.42\textwidth}{!} 
				{
				$\left\lbrace\left\lbrace x^i_j, y^i_j \right\rbrace^{M-1}_{j=0}\right\rbrace^{N}_{i=1},j\in \left\lbrace 0,1,2,...,M-1\right\rbrace,i\in \left\lbrace 1,2,3,...,N \right\rbrace$
				}
			\end{equation}
			where $x^i_j$ is a raw pixel at location \textit{j} of example \textit{i}, $y^i_j$ is a discrete variable indicating the true class of the raw pixel at location \textit{j} of example \textit{i}. A binary cross entropy loss function is used to compare the true label $y^i_j$ and the predicted value. The binary cross entropy loss function of example \textit{i} is defined by:
			\begin{equation}
				L_i=\frac{-1}{M}\sum_{j=1}^{M}[y^i_jlog(p^i_j)+(1-y^i_j)log(1-p^i_j)]
			\end{equation}
			where $p^i_j$ is the predicted probability value of the pixel at location \textit{j} of example \textit{i}. Note that we do not associate any labels to the unknown regions like the non-Region of Interest and the boundary of objects in our loss computations during training. We observed that this deliberate avoidance makes our network more confident in pixel prediction.
			
			We train our network using RMSProp optimizer with a batch size of 1, setting rho to 0.9 and epsilon to 1e-08. In fine-tuning, we use a small learning rate, 1e-4, since we do not want to change the existing weights, which are initially good for feature representations. We reduce the learning rate by a factor of 10 when minimum validation loss stops improving for 6 epochs. We train using 60 epochs for 50 training examples case and 50 epochs for 200 training examples case. Model checkpoint is used to save the best model when validation loss decreases; in practice, we let model checkpoint select the best model for us. We set the L2 regularization strength to 5e-4 in our experiments.
			
			One important thing to note is that due to the nature of video sequences, training frames by reading and feeding to the network in a sequential order may cause a bias in the learned weights, since many frames in a row contain very similar content. In practice, to prevent this issue, we perform random shuffling in two phases: random shuffling of training frames before splitting the frames into training/validation set and random shuffling before each epoch in the training process. We observed that our network benefit from these processes and converge faster. After shuffling, we perform a validation split of 20\%, hence 80\% of the training examples are used to train the model.
			
			There are around 8.2M parameters in total, which contains 6.5M trainable parameters and about 1.7M non-trainable parameters in our network. About 93\% of total parameters come from VGG-16 Net and other 7\% comes from TCNN part; the parameters in TCNN is significantly less due to the dimension reduction by 1x1 transposed convolutional layers. As described in Section \ref{sec:3.1.1}, we penalize the loss if a foreground pixel is classified as a background pixel during training process. This helps improving the network performance by some margins. Note that we do not perform any kind of input normalizations, including the mean subtractions during the training.
			
			As illustrated in Fig. \ref{fig:fig1} and described in Section \ref{sec:3.2}, three different scales of the same image are constructed using Gaussian filtering. These scaled-images pass through the network simultaneously to produce feature maps in three different scales. Each feature map is upsampled and concatenated to form the input of the TCNN to produce the probability mask in the original image size.
			
\section{Experiments}
\label{sec:4}
	 Since the outputs of our network is a probability mask that contains values between 0 and 1 for each pixel, we use a threshold to convert these probabilities to binary masks. We set the mask corresponding to a pixel to 1 if it's probability exceeds a given threshold. Fig. \ref{fig:fig4} illustrates our network's classification performance for a set of different threshold values; for the experiments that we use 200 frames, it shows that a threshold of 0.9 gives the best average F-Measure across 11 categories. This high probability indicates that our method is extremely confident with its predictions overall. For 50 frames case, the confidence level decreases slightly, i.e. a threshold of 0.7 gives the best score. To fix the threshold to a specific value for both experiments, we choose a fixed threshold value of 0.8 in all our experiments. Note that no post-processing is applied after thresholding such as conditional random field (CRF) or utilization of any other graphical models etc. to ensure the consistency of final segmentation mask in our implementation. 
	 \begin{center}
	 	\includegraphics[width=0.9\columnwidth, center]{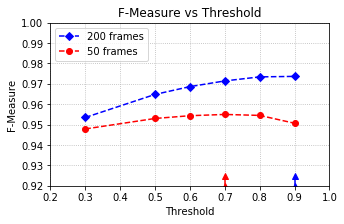}
	 	\captionsetup{width=0.9\columnwidth}
	 	\captionof{figure}{An illustration of average F-Measure (on test set across 11 categories) versus different thresholds. Arrows indicate the highest F-Measure value.}
	 	\label{fig:fig4}
	 \end{center}
	\subsection{Evaluation Metrics}
	\label{sec:4.1}
		In this section, we briefly discuss three different metrics for the model performance evaluation: F-Measure, MCC and Percentage of Wrong Classifications (PWC). Given true positive (\textit{TP}), false positive (\textit{FP}), false negative (\textit{FN}) and without incorporating true negative (\textit{TN}); the harmonic mean of the precision and recall, F-Measure, is defined by:
		\begin{equation}
			F-Measure = \frac{2\times precision\times recall}{precision + recall}
		\end{equation}
		where $precision=\frac{TP}{TP + FP}$, $recall=\frac{TP}{TP + FN}$. F-Measure is in the range of [0,1] where a value of 1 indicates that the predicted mask totally agrees with its ground-truth, on the other hand, a value of 0, indicates disagreement. By incorporating \textit{TN}, PWC is defined by:
		\begin{equation}
			PWC = \frac{100\times (FP + FN)}{TP + FP + TN + FN}
		\end{equation}
		However, as stated previously, F-Measure is sensitive to imbalanced classes; consider that there is no foreground pixel in a frame, in this case, F-Measure will be zero although our model correctly classifies all background pixels as the background. To overcome this issue, MCC metric is used in performance measures due to its stability to imbalanced classes problem. MCC metric is defined in terms of \textit{TP, FP, FN} and \textit{TN} by:
		\begin{equation}
			\resizebox{0.42\textwidth}{!} 
			{
				$MCC = \frac{(TP\times TN)-(FP\times FN)}{\sqrt{(TP+FP)(TP+FN)(TN+FN)(TN+FP)}}$
			}
		\end{equation}
		where its value is in the range of [-1, +1], where a value of +1 indicates that the predicted mask totally agrees with its ground-truth, and a value of -1 indicates disagreement.
		\begin{table*}[!th]
			\renewcommand{\arraystretch}{1.5} 
			\setlength{\tabcolsep}{4pt}
			\centering	
			\caption{The results are obtained by manually and randomly selecting, 50 and 200 frames from CDnet2014 dataset. Each row shows the average results of each category. The last row shows the average results across 11 categories. Note that only the test frames are included in the reported performances.}
			\label{table:2}
			\tiny \begin{tabular}{*c^c^c^c^c^c^c^c^c^c^c^c^c >{\bfseries}c>{\bfseries}c}
				\hline
				\multirow{2}{*}{Category} & \multicolumn{2}{c}{Recall} & \multicolumn{2}{c}{Specificity} & \multicolumn{2}{c}{FPR} & \multicolumn{2}{c}{FNR} & \multicolumn{2}{c}{PWC} & \multicolumn{2}{c}{Precision} & \multicolumn{2}{c}{\textbf{F-Measure}}\\\cline{2-15}
				&50f&200f&50f&200f&50f&200f&50f&200f&50f&200f&50f&200f&50f&200f\\
				\hline
				baseline&0.9887&0.9951&0.9999&1.0000&0.00009&0.00003&0.0113&0.0049&0.0405&0.0152&0.9964&0.9986&0.9926&0.9968\\
				cam. jitter&0.9702&0.9878&0.9997&0.9998&0.00034&0.00018&0.0298&0.0122&0.1696&0.0605&0.9906&0.9950&0.9801&0.9914\\
				bad weath.&0.9484&0.9759&0.9997&1.0000&0.00029&0.00024&0.0516&0.0241&0.1180&0.0494&0.9805&0.9755&0.9636&0.9757\\
				dyna. bg.&0.9826&0.9906&0.9999&1.0000&0.00007&0.00003&0.0174&0.0094&0.0208&0.0071&0.9883&0.9826&0.9854&0.9865\\
				inter. obj.&0.9670&0.9889&0.9995&0.9998&0.00048&0.00020&0.0330&0.0111&0.1295&0.0823&0.9833&0.9956&0.9749&0.9922\\
				low f.rate&0.8374&0.8990&0.9996&0.9999&0.00035&0.00010&0.1626&0.1010&0.1000&0.0336&0.8049&0.8687&0.8164&0.8816\\
				night vid.&0.8817&0.9606&0.9993&0.9997&0.00069&0.00033&0.1183&0.0394&0.2740&0.0992&0.9671&0.9788&0.9216&0.9696\\
				PTZ&0.9350&0.9755&0.9999&0.9999&0.00013&0.00005&0.0650&0.0245&0.0523&0.0164&0.9779&0.9417&0.9557&0.9567\\
				shadow&0.9839&0.9922&0.9994&0.9999&0.00063&0.00011&0.0161&0.0078&0.1211&0.0374&0.9768&0.9966&0.9800&0.9944\\
				thermal&0.9598&0.9871&0.9995&0.9998&0.00052&0.00016&0.0402&0.0129&0.2042&0.0683&0.9859&0.9944&0.9725&0.9907\\
				turbulence&0.9443&0.9675&1.0000&0.9999&0.00015&0.00009&0.0557&0.0325&0.0426&0.0264&0.9704&0.9772&0.9571&0.9722\\
				\hline
				\rowstyle{\bfseries}Overall&0.9454&0.9746&1.0000&0.9999&0.00034&0.00014&0.0546&0.0254&0.1156&0.0451&0.9656&0.9732&0.9545&0.9734 \\
				\hline
			\end{tabular}
		\end{table*}
	\subsection{CDnet2014 Dataset}
		We utilize CDnet2014 dataset \cite{wang2014cdnet} in our experiment. CDnet2014 dataset contains 11 categories: baseline, camera jitter, bad weather, dynamic background, intermittent object motion, low frame rate, night videos, PTZ (pan-ning-tilting-zooming), shadow, thermal and turbulence. Each category contains from 4 to 6 sequences. Totally, there are 53 different video sequences. Spatial resolutions of video frames vary from 320x240 to 720x576 pixels. Moreover, a video sequence may contain from 900 to 7000 frames. Almost all video sequences contain different challenging scenarios which make this dataset appropriate for measuring the robustness of a model in each such cases.
		
	\subsection{Results and Discussion}
		We use 50 and 200 frames randomly as the training frames in our - two sets of - experiments. We will refer to the experiments in which we use only 50 training frames as 50-frame experiments; similarly, we will refer to the experiments in which we use only 200 training frames as 200-frame experiments from now on.
		
		We follow the same training frame selection strategy, randomly manual selection, as described in \cite{WANG201766}. Firstly, we perform experiments using a set of frames that we selected manually and report test results by considering only the range of the frames that contain the ground truth labels. The results of these experiments are depicted in Table \ref{table:2}. Note that, these values are computed using only the test frames, i.e. the training frames are excluded in the performance evaluation. With this setting, we get an overall F-Measure of 0.9545 with 50-frame experiments and 0.9734 with 200-frame experiments.
		
		As can be seen in Table \ref{table:2}, in the 200-frame experiments, our network provides high accuracy in foreground segmentation. In the \textit{baseline} category, we obtain the highest average F-Measure compared to the other categories. By reducing the number of training examples to 50 frames, F-Measure decreases by some margins. Especially, in \textit{lowFrameRate} category, F-Measure decreases by 6.5\% compared to the model with 200 training examples. However, we still obtain acceptable results with an average overall F-Measure of 0.9545 across 11 categories, which outperforms the current state-of-the-art methods. This shows that our method works robustly in many challenging foreground objects segmentation domains.
		\begin{table*}[!t]	
			\renewcommand{\arraystretch}{1.5}
			\centering
			\captionsetup{skip=1pt}
			\caption{A category-wise comparison of F-Measure across 11 categories among 6 methods. Each row shows the results for each method. Each column shows the average results in each category. Note that we consider all the frames in the ground-truths of CDnet2014 dataset.}
			\label{table:3}
			\tiny \begin{tabular}{*c^c^c^c^c^c^c^c^c^c^c^c}
				\hline
				\multirow{2}{*}{Methods}&\multicolumn{11}{c}{\textbf{F-Measure (category-wise)}} \\ \cline{2-12}
				&baseline&cam.jitter&bad.weat&dyna.bg&int.obj.m.&low f.rate&night vid.&PTZ&shadow&thermal&turbul.\\
				\hline
				\rowstyle{\bfseries}FgSegNet&0.9975&0.9945&0.9838&0.9939&0.9933&0.9558&0.9779&0.9893&0.9954&0.9923&0.9776 \\
				Cascade\cite{WANG201766}&0.9786&0.9758&0.9451&0.9658&0.8505&0.8804&0.8926&0.9344&0.9593&0.8958&0.9215 \\
				DeepBS\cite{babaee2017deep}&0.9580&0.8990&0.8647&0.8761&0.6097&0.5900&0.6359&0.3306&0.9304&0.7583&0.8993 \\
				IUTIS-5\cite{bianco2017far}&0.9567&0.8332&0.8289&0.8902&0.7296&0.7911&0.5132&0.4703&0.9084&0.8303&0.8507 \\
				PAWCS\cite{st2015self}&0.9397&0.8137&0.8059&0.8938&0.7764&0.6433&0.4171&0.4450&0.8934&0.8324&0.7667 \\
				SuBSENSE\cite{st2015subsense}&0.9503&0.8152&0.8594&0.8177&0.6569&0.6594&0.4918&0.3894&0.8986&0.8171&0.8423 \\
				\hline
				\vspace{0.1pt}
			\end{tabular}
		\end{table*}
		\begin{table*}[!]
			\renewcommand{\arraystretch}{1.5}
			\centering
			\captionsetup{skip=1pt}
			\caption{A category-wise comparison of MCC across 11 categories among 6 methods. Note that we consider all the frames in the ground-truths of CDnet2014 dataset.}
			\label{table:4}
			\tiny \begin{tabular}{*c^c^c^c^c^c^c^c^c^c^c^c}
				\hline
				\multirow{2}{*}{Methods}&\multicolumn{11}{c}{\textbf{MCC (category-wise)}} \\ \cline{2-12}
				&baseline&cam.jitter&bad.weat&dyna.bg&int.obj.m.&low f.rate&night vid.&PTZ&shadow&thermal&turbul.\\
				\hline
				\rowstyle{\bfseries}FgSegNet&0.9975&0.9942&0.9836&0.9938&0.9929&0.9557&0.9774&0.9892&0.9952&0.9920&0.9775 \\
				Cascade\cite{WANG201766}&0.9780&0.9748&0.9443&0.9658&0.8591&0.8798&0.8911&0.9349&0.9577&0.8932&0.9230 \\
				DeepBS\cite{babaee2017deep}&0.9571&0.8976&0.8718&0.8777&0.6371&0.6061&0.6617&0.3701&0.9284&0.7609&0.9024 \\
				IUTIS-5\cite{bianco2017far}&0.9553&0.8274&0.8333&0.8932&0.7406&0.7943&0.5182&0.5002&0.9060&0.8328&0.8584 \\
				PAWCS\cite{st2015self}&0.9375&0.8121&0.8120&0.8936&0.7737&0.6573&0.4327&0.4911&0.8875&0.8282&0.7857 \\
				SuBSENSE\cite{st2015subsense}&0.9487&0.8080&0.8596&0.8240&0.6738&0.6860&0.4998&0.4442&0.8958&0.8098&0.8448 \\
				\hline
				\vspace{0.1pt}
			\end{tabular}
		\end{table*}
	
		\begin{table*}[!]
			\renewcommand{\arraystretch}{1.5}
			\centering
			\captionsetup{skip=1pt}
			\caption{Average results across 11 categories for each methods. Note that we consider all the frames in the ground-truths of CDnet2014 dataset.}
			\label{table:5}
			\scriptsize \begin{tabular}{*c^c^c^c^c^c}
				\hline
				\multirow{2}{*}{Methods}&\multicolumn{5}{c}{\textbf{Overall}} \\ \cline{2-6}
				&Precision&Recall&PWC&F-Measure&MCC \\
				\hline
				\rowstyle{\bfseries}FgSegNet&0.9889&0.9841&0.0426&0.9865&0.9863 \\
				Cascade\cite{WANG201766}&0.9048&0.9584&0.3882&0.9272&0.9274 \\
				DeepBS\cite{babaee2017deep}&0.8401&0.7650&1.8699&0.7593&0.7701 \\
				IUTIS-5\cite{bianco2017far}&0.8105&0.7972&1.0863&0.7820&0.7872 \\
				PAWCS\cite{st2015self}&0.7841&0.7724&1.1196&0.7477&0.7556 \\
				SuBSENSE\cite{st2015subsense}&0.7522&0.8144&1.5869&0.7453&0.7540 \\
				\hline
			\end{tabular}
		\end{table*}
	
		\begin{table*}[!]
			\setlength{\tabcolsep}{1.6pt}
			\renewcommand{\arraystretch}{1.5}
			\centering
			\captionsetup{skip=1pt}
			\caption{A comparison between our method and the current best method in CDnet2014 benchmark. Note that we evaluate these scores by excluding the training frames.}
			\label{table:6}
			\tiny \begin{tabular}{*c^c^c^c^c^c^c^c^c^c^c^c^>{\bfseries}c|^c|^c}
				\hline
				\multirow{2}{*}{Methods}&\multicolumn{12}{c|}{\textbf{F-Measure}}&\textbf{MCC}& \multirow{2}{*}{\textbf{Seg./Train. Speed}} \\ \cline{2-14}
				&baseline&cam.Jitter&badWea.&dyna.bg&int.obj&lowF.rate&nightVid.&PTZ&shadow&thermal&turbul.&Overall&\textbf{Overall}&\\
				\hline
				\rowstyle{\bfseries}FgSegNet&0.9968&0.9914&0.9757&0.9865&0.9922&0.8816&0.9696&0.9567&0.9944&0.9907&0.9722&0.9734&0.9734&$\sim$18fps/23.7min \\
				Cascade\cite{WANG201766}&0.9779&0.9687&0.9421&0.6515&0.8225&0.7373&0.8882&0.7052&0.9548&0.8785&0.9190&0.8587&\textbf{0.8600}&$\sim$13fps/35min \\
				\hline
				\vspace{0.1pt}
			\end{tabular}
		\end{table*}
		\begin{table*}[!]
			\renewcommand{\arraystretch}{1.5}
			\centering
			\captionsetup{skip=1pt}
			\caption{A comparison between our method and the current top methods in CDnet2014 benchmark. Note that these results obtained from CDnet 2014 challenge website\protect\textcolor{magenta}{\textsuperscript{\ref{1}}}.}
			\label{table:7}
			\scriptsize \begin{tabular}{*c^c^c^c^c }
				\hline
				\multirow{2}{*}{Methods}&\multicolumn{4}{c}{\textbf{Overall}} \\ \cline{2-5}
				&Avg. Precision & Avg. Recall & Avg. PWC & Avg. F-Measure \\
				\hline
				\rowstyle{\bfseries}FgSegNet&0.9758&0.9836&0.0559&0.9770 \\
				Cascade\cite{WANG201766}&0.8997&0.9506&0.4052&0.9209 \\
				DeepBS\cite{babaee2017deep}&0.8332&0.7545&1.9920&0.7458 \\
				IUTIS-5\cite{bianco2017far}&0.8087&0.7849&1.1986&0.7717 \\
				PAWCS\cite{st2015self}&0.7857&0.7718&1.1992&0.7403 \\
				SuBSENSE\cite{st2015subsense}&0.7509&0.8124&1.6780&0.7408 \\
				\hline
			\end{tabular}
		\end{table*}
		
		In order to compare our results with the previous methods, we need to consider all the frames, i.e. both training and testing frames, in performance evaluations, since previous methods included all frames. The F-Measure and MCC performances of different methods for each category are provided in Table \ref{table:3} and Table \ref{table:4}, respectively. The overall performances across 11 categories are depicted in Table \ref{table:5}. Deep learning based-methods perform better in very challenging categories such as \textit{nightVideo} and \textit{PTZ}; yet, they do not perform well in \textit{lowFrameRate} category. However, most of the deep learning methods still perform better than the conventional approaches by large margins.
		\begin{figure*}[h!]
			\includegraphics[width=\textwidth, height=0.93\textheight]{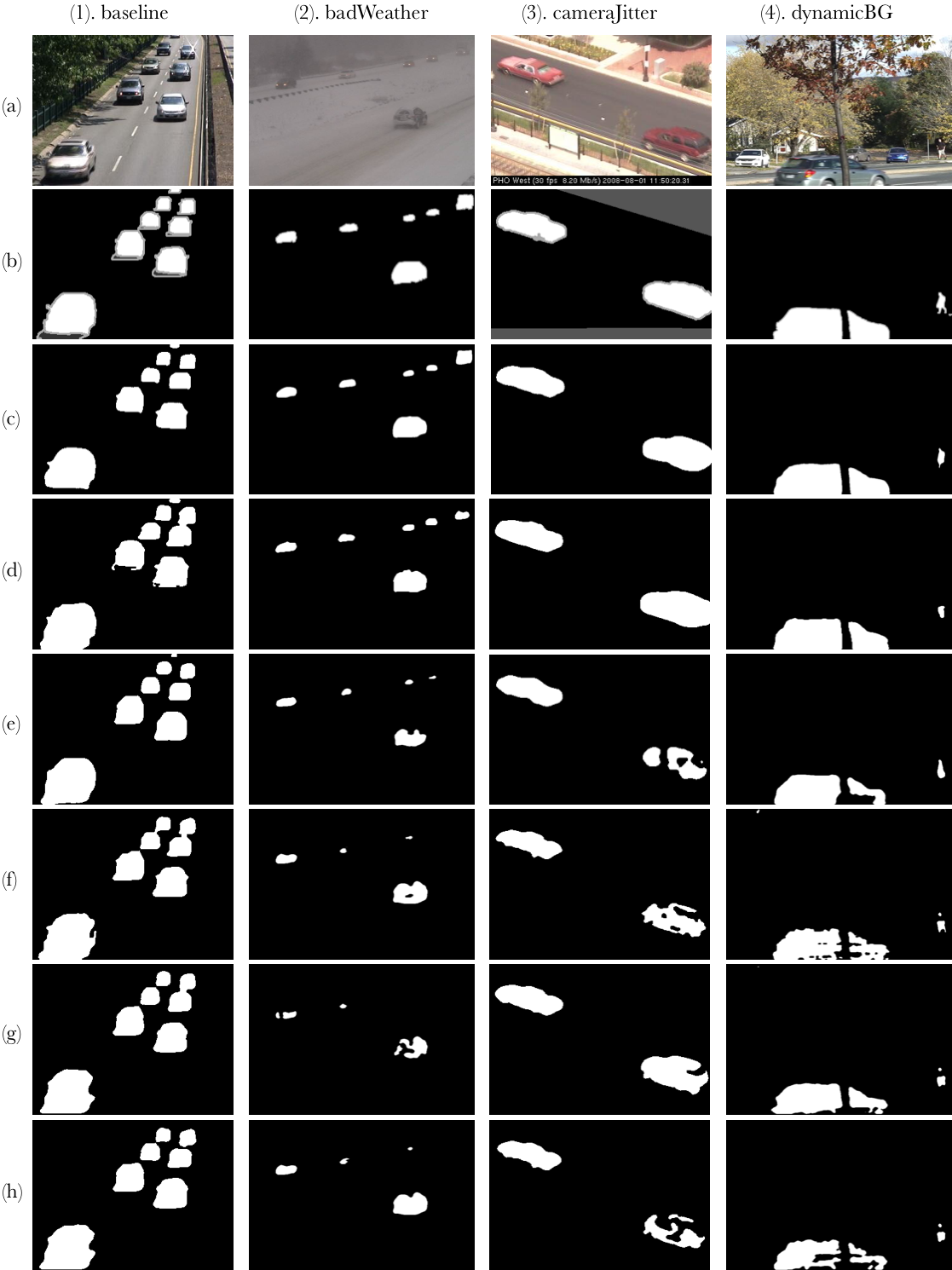}
			\caption{Results obtained from a selected scene in each category. (a) shows raw images, (b) shows  the ground-truths, (c) shows the results obtained from our method. (d), (e), (f), (g) and (h) show the results obtained from Cascade, DeepBS, IUTIS-5, PAWCS and SuBSENSE, respectively.}
			\centering
			\label{fig:fig5}
		\end{figure*}
		\begin{figure*}[!]
			\includegraphics[width=\textwidth, height=0.93\textheight, center]{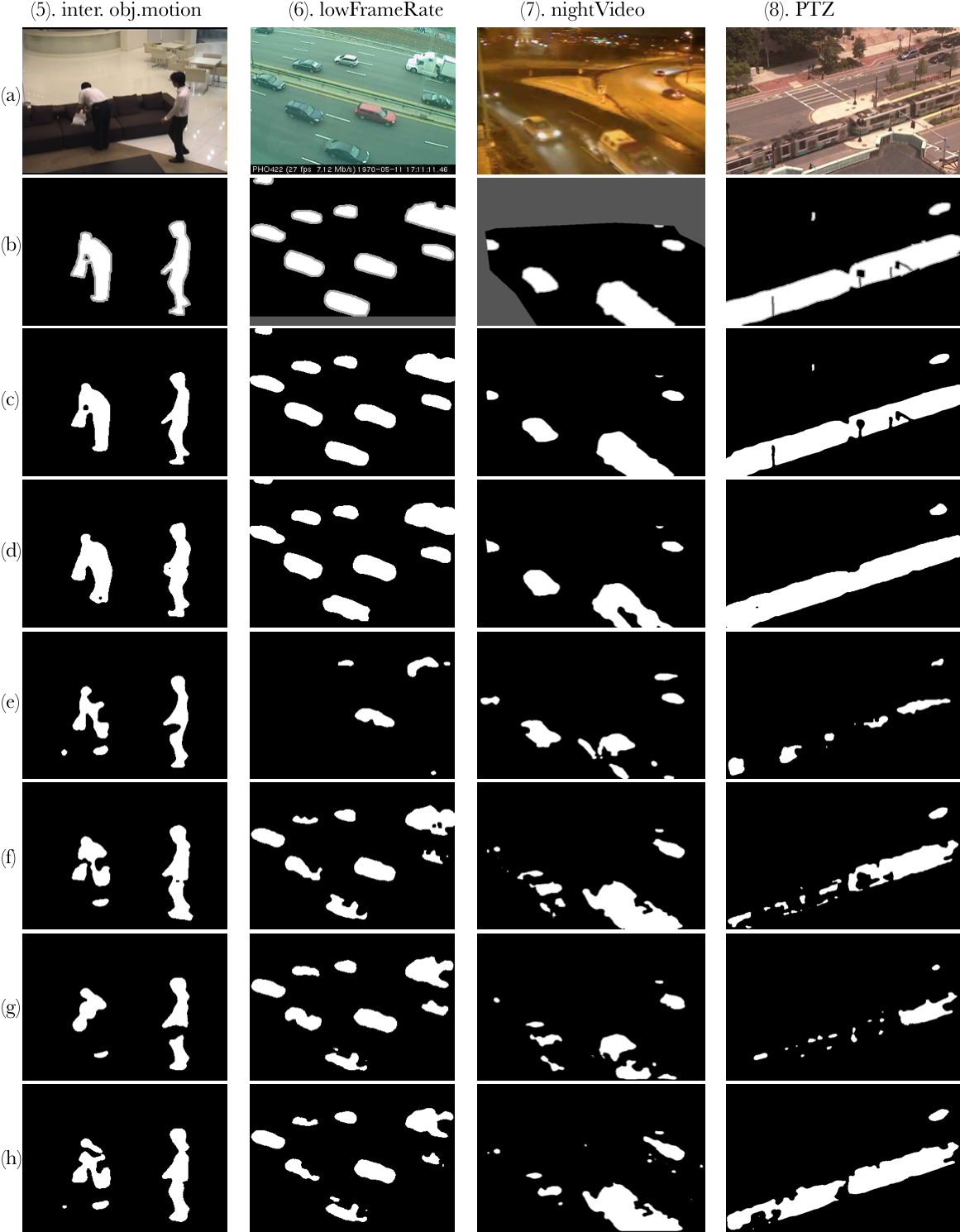}
			\caption{Results obtained from a selected scene in each category. (a) shows raw images, (b) shows  the ground-truths, (c) shows the results obtained from our method. (d), (e), (f), (g) and (h) show the results obtained from Cascade, DeepBS, IUTIS-5, PAWCS and SuBSENSE, respectively.}
			\centering
			\label{fig:fig6}
		\end{figure*}
		
		We also provide a comparison of the scores by excluding the training samples in Table \ref{table:6}. Here, we only pick the current best method in the CDnet 2014 challenge \cite{WANG201766} to compare with our method. Since the frame-level foreground masks and training frames are publicly available for this work, we are able to make fair comparisons.
		
		In addition to these experiments, we also perform additional experiments by training our model using the training frames provided by the authors in \cite{WANG201766}, by only adjusting four scenes, and report the results that we obtained from CDnet 2014 challenge in Table \ref{table:7}. These results show the performance of the methods with the dataset as a whole, i.e. including additional frames where ground truth values are not shared with the public dataset. Our method outperforms the current state of the art methods, hence is ranked as number one in the performance evaluations of the CDnet 2014 challenge web framework as well.
		
\begin{figure}[!ht]
			\setlength{\belowcaptionskip}{5pt}
			\includegraphics[width=0.92\columnwidth, height=0.55\textheight, center]{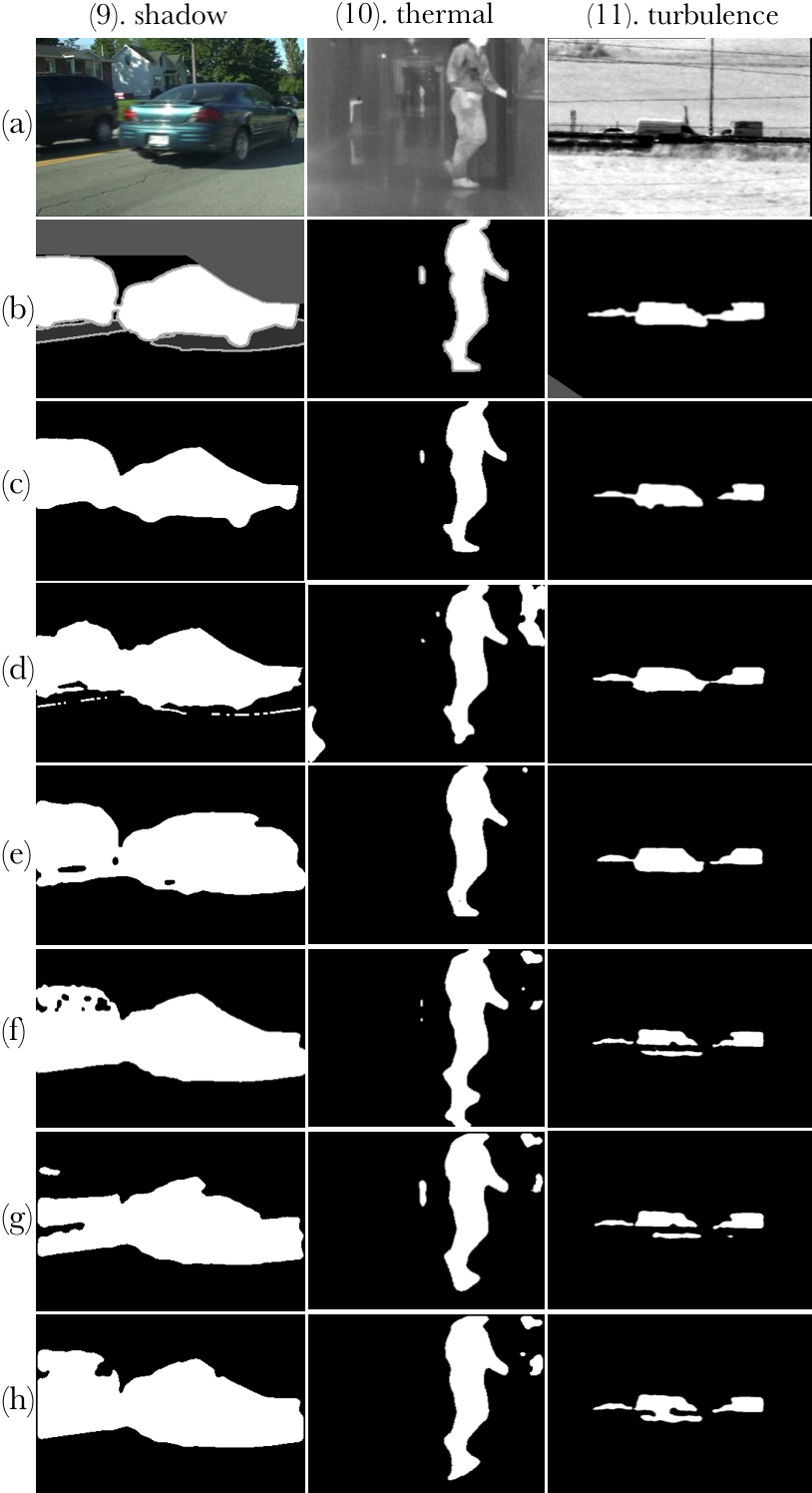}
			\captionsetup{width=0.9\columnwidth}
			\caption{(a) raw images, (b) ground-truths, (c) results obtained from our method. (d), (e), (f), (g) and (h) show the results obtained from Cascade, DeepBS, IUTIS-5, PAWCS and SuBSENSE, respectively.}
			\centering
			\label{fig:fig7}
		\end{figure}

		We provide some exemplary results in Fig. \ref{fig:fig5}, Fig. \ref{fig:fig6} and Fig. \ref{fig:fig7} that illustrate the foreground masks that are estimated by 6 different methods. Due to space limitations, we pick a scene from each category randomly in these figures. As can be seen from these figures, our model can estimate more accurate object boundaries even when the foreground objects are very small and ambiguous. It can also eliminate dynamic backgrounds and shadows completely. Furthermore, our model is also robust against large camera motions, as can be seen in \textit{cameraJitter} and \textit{PTZ} categories that contain various camera movements; we obtained more than 0.95 in the F-Measure in both categories.
		
\begin{figure}[!h]
			\setlength{\belowcaptionskip}{-8pt}
			\includegraphics[width=0.9\columnwidth, height=0.26\textheight, center]{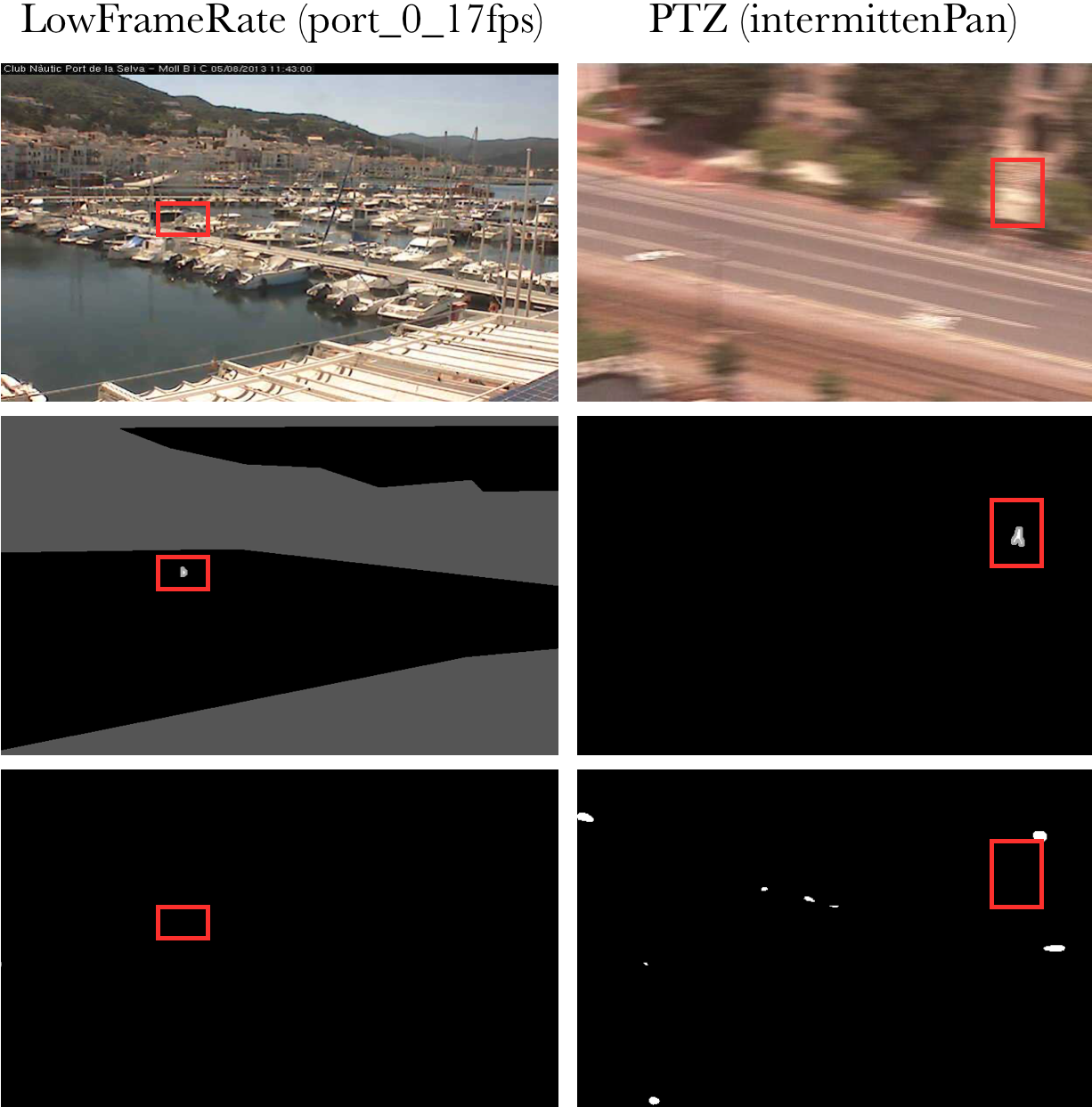}
			\captionsetup{width=0.9\columnwidth}
			\caption{Sample scenes that our model perform poorly. First row shows raw images, second row shows the ground-truths. Third row shows our segmentation results.}
			\centering
			\label{fig:fig8}
		\end{figure}

		Our method performs poorly for \textit{lowFrameRate} category, when we compare the F-Measures with the other categories. This low performance is primarily due to the challenging content in a scene where there are extremely small foreground objects displayed in low frame rates. It is even hard for a human observer to spot the spatial positions of the objects. An example is provided in Fig. \ref{fig:fig8}; the first column explains the challenge. The second low performance is observed in \textit{PTZ} (pan-tilt-zoom) category (Fig. \ref{fig:fig8}, second column), where camera continuously pan-tilt-zoom around the scene and making blurry scenes; as can clearly be seen from the scene, when camera starts panning, the foreground object (a human wearing a white shirt who is walking along the sidewalk, in this case) is blended with the background completely. Even for a human observer, it is hard to distinguish whether that region contains foreground object or not. In this category, our model produces more false positives. However, the average score of \textit{PTZ} category is still acceptable compared to other categories.
	\subsection{Processing Speed}	
		As stated previously, we use Keras framework with Tensorflow backend and boost our implementation process using NVIDIA GTX 970 GPU. Considering a video sequence which has 1700 frames with spatial size of 320x240, with a 200-frame training, it takes around 23.7 minutes to train for 50 epochs. For testing with the remaining 1500 frames, it takes around 1.39 minutes to segment; which means that our network is capable of segmenting about 18 frames per second. Hence, compared to the best previous method, our method is faster in terms of training and segmentation speed (Table \ref{table:6}, right-most column).
\section{Conclusion}
\label{sec:5}
	In this work we demonstrate a robust and flexible encoder-decoder type network model that produces high accuracy segmentation masks for a variety of challenging scenes. The model is trained end-to-end in a supervised manner; it is fed with multi-scaled raw RGB images. It does not require any post-processing on the generated segmentations masks. We adapt VGG-16 Net by modifying some of the higher layers, keeping the lower layers unchanged under a triplet network configuration. We embed transposed convolutional layers on the top of the network to upscale feature maps back to the image space. We utilize many bottleneck layers (1x1 transposed convolutions) in the transposed convolutional layers reduce the dept dimensions.
	
	In the context of this research, we tried different training examples selection strategies; we suggest frame selection in such a way to increase variety of the scenes from different parts of the video sequence, whether it contains foreground objects or not. Moreover, we provide a solution to imbalanced data problem in this domain; we alleviate this problem by introducing weight penalization when foreground pixels are classified as background pixels.
	
	Our method is robust against various difficult situations such as illumination changes, background or camera motion, camouflage effect, shadow etc. It can be deployed in both indoor or outdoor scenes. We also show that our method outperforms all the existing methods including previous best deep learning based-method.
	
	Our model is learning foreground objects by isolated frames, i.e. the time sequence is not considered during training. As a future work, we plan to redesign our network to learn from temporal data by operating 3D convolutional networks with different fusion techniques.
	
\section*{Acknowledgements}
	We would like to thank CDnet2014 benchmark \cite{wang2014cdnet} for making the segmentation masks of all methods publicly available, which allowed us to perform different types of comparisons. And we also thank the authors in \cite{WANG201766} who made their training frames publicly available for follow-up researchers.

\bibliographystyle{ieeetr}
\footnotesize \bibliography{fgsegnet}
\end{document}